\documentclass{article}
\usepackage{ijcai26}

\usepackage{times}
\usepackage{soul}
\usepackage{url}
\usepackage[hidelinks]{hyperref}
\usepackage[utf8]{inputenc}
\usepackage[small]{caption}
\usepackage{graphicx}
\usepackage{amsmath,amssymb}
\usepackage{amsthm}
\usepackage{booktabs}
\usepackage{algorithm}
\usepackage{algpseudocode}
\usepackage[switch]{lineno}
\usepackage{enumerate}
\usepackage{enumitem}
\usepackage{bm}
\usepackage{multirow}
\usepackage{makecell}
\usepackage{subcaption}

\usepackage{xcolor}

\newtheorem{theorem}{Theorem}

\title{Neural Probe–Based Hallucination Detection for Large Language Models}

\author{
	Shize Liang\textsuperscript{1}
	\and
	Hongzhi Wang\textsuperscript{2,*}
	\affiliations
	$^1$Faculty of Computing, Harbin Institute of Technology\\
	$^2$Faculty of Computing, Harbin Institute of Technology\\
	\emails
	2023111781@stu.hit.edu.cn,
	wangzh@hit.edu.cn
}

\begin{document}
	
	\maketitle
	
	\begin{abstract}
		Large language models(LLMs) excel at text generation and knowledge question-answering tasks, but they are prone to generating hallucinated content, severely limiting their application in high-risk domains. Current hallucination detection methods based on uncertainty estimation and external knowledge retrieval suffer from the limitation that they still produce erroneous content at high confidence levels and rely heavily on retrieval efficiency and knowledge coverage. In contrast, probe methods that leverage the model's hidden-layer states offer real-time and lightweight advantages. However, traditional linear probes struggle to capture nonlinear structures in deep semantic spaces.
		To overcome these limitations, we propose a neural network-based framework for token-level hallucination detection. By freezing language model parameters, we employ lightweight MLP probes to perform nonlinear modeling of high-level hidden states. A multi-objective joint loss function is designed to enhance detection stability and semantic disambiguity. Additionally, we establish a layer position–probe performance response model, using Bayesian optimization to automatically search for optimal probe insertion layers and achieve superior training results.
		Experimental results on \emph{LongFact}, \emph{HealthBench}, and \emph{TriviaQA} demonstrate that MLP probes significantly outperform state-of-the-art methods in accuracy, recall, and detection capability under low false-positive conditions.
		
	\end{abstract}

\section{Introduction}
	
	Large Language Models demonstrate exceptional capabilities in tasks such as text generation, knowledge-based question answering, reasoning, and summarization~\cite{matarazzo2025surveylargelanguagemodels}. They are widely deployed across high-stakes domains such as healthcare, finance, and law~\cite{Wang2023ClinicalGPT}. However, despite their prowess in natural language understanding and generation, LLMs frequently produce "hallucinations", the output that is factually inaccurate or lacks an evidentiary basis. In high-stakes domains, hallucinations can lead to severe consequences, such as misleading medical diagnoses, erroneous legal citations, or the dissemination of fake news~\cite{Tang2023MedicalSummarization}. Consequently, developing accurate and efficient mechanisms for detecting and mitigating hallucinations has become a critical research direction for enhancing the reliability and interpretability of LLMs.
	
	Existing hallucination detection methods mainly fall into three categories. Uncertainty-based approaches estimate confidence using perplexity, semantic entropy, or sampling consistency~\cite{kossen2024semanticentropyprobesrobust}. These models often remain confidently wrong. Retrieval-based methods verify content using structured knowledge bases~\cite{abdi-etal-2025-hallurag}, yet depend heavily on retrieval speed and knowledge coverage, limiting real-time use. A third direction analyzes internal model representations~\cite{obeso2025realtimedetectionhallucinatedentities}. While lightweight probes can detect hallucinations in real time, traditional linear probes capture only shallow local patterns and miss the nonlinear semantics encoded in deep hidden states.
	
	\begin{figure*}[h]
		\centering
		\includegraphics[width=0.95\linewidth]{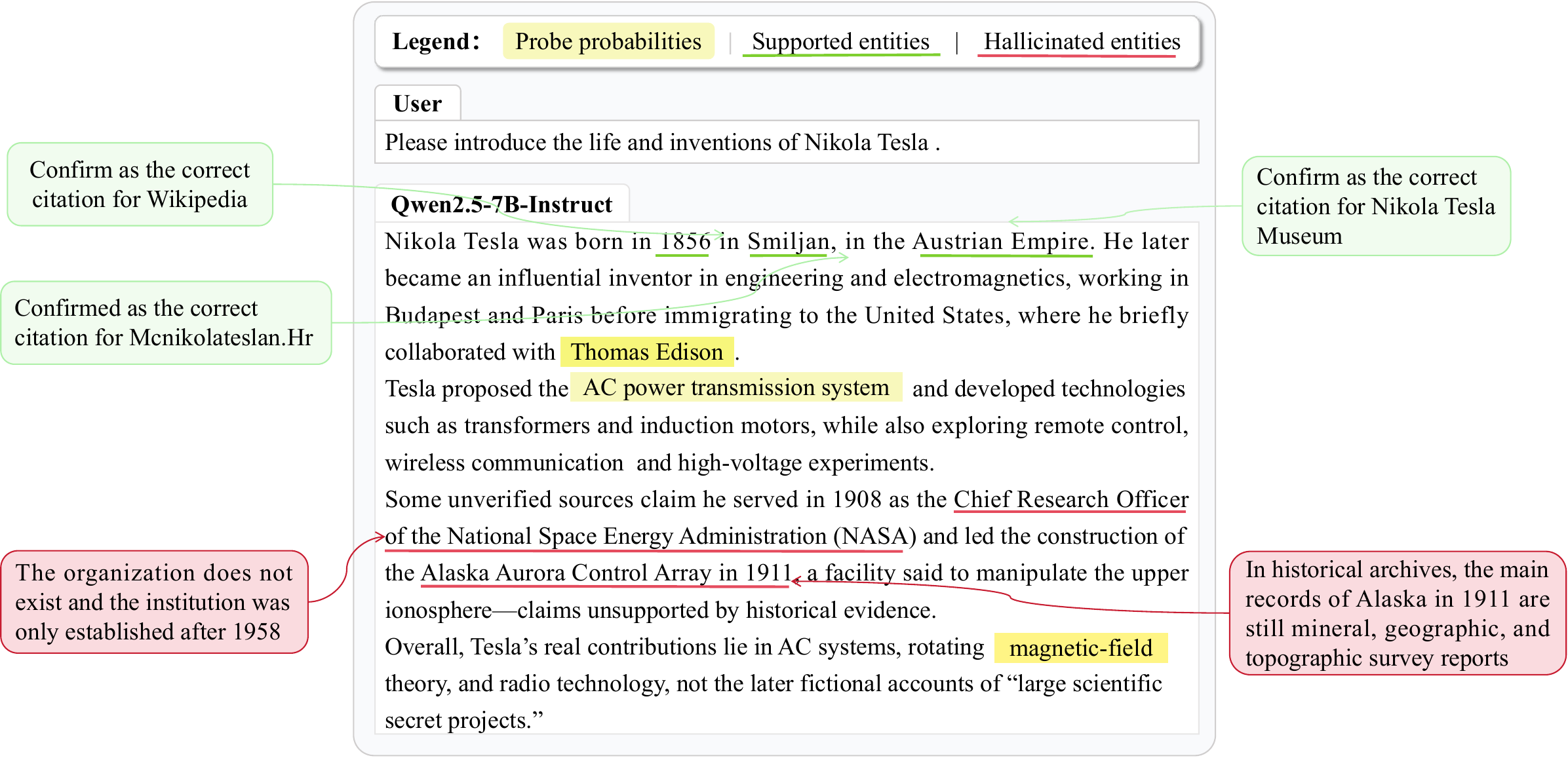}
		\caption{Annotated examples of hallucination detection text, where each token's hallucination detection probe score is highlighted in yellow. The intensity of the color reflects the magnitude of the score: green indicates supported entities, while red denotes entities flagged as hallucinations.}
		\label{conversation}
	\end{figure*}
	
	In hallucination detection for large language models, real-time and precise identification of generated content authenticity is crucial for enhancing model reliability. Current research primarily focuses on statement-level hallucinations, with limited attention given to more complex entity-level hallucinations (e.g., fabricated names, dates, citations) that are harder to annotate. However, entities possess clearer token boundaries than assertions and support real-time verification, whereas assertions often require extraction after the model generates complete sentences—a process that disrupts token alignment and delays system responses. To address this, this paper proposes a token-level hallucination detection framework based on neural network probes. This framework focuses on entity-level hallucinations. While keeping pre-trained LLM parameters frozen, it employs lightweight deep neural networks as probes to hierarchically model the high-level semantic space within the model. Combined with a multi-objective joint loss, it achieves stable optimization of semantic separability. Compared to traditional linear probes, our approach effectively captures complex nonlinear hallucination patterns at the token-level semantic scale, significantly improving hallucination detection performance with reduced training time.
	
	The main contributions of this work are summarized as follows:
	
	\textbullet \ We propose a neural network probe-based token-level hallucination detection framework that enables real-time and efficient use of internal LLM representations.
	
	\textbullet \ To enhance detection stability and semantic discrimination, we design a multi-objective joint loss that combines token-level focal loss, soft span aggregation, sparsity regularization, and KL-divergence constraints.
	
	\textbullet \ We establish a layer-position–probe-performance model and introduce a Bayesian optimization–based strategy that replaces heuristic probe layer selection.
	
	\textbullet \ Experiments on \emph{LongFact} and \emph{TriviaQA} demonstrate the efficacy of our hierarchical semantic modeling strategy by showing that our method outperforms strong baselines in both recall and accuracy, with precision on \emph{TriviaQA} improving by more than 270\%.

\section{Related Work}
	
	\subsection{Hallucination Detection Mechanism}	
	Existing hallucination detection methods focus on various implementation approaches. ~\cite{Ji_2023} utilized perplexity, semantic entropy, and sampling consistency metrics to measure a model's confidence in its output. Building upon this, ~\cite{kossen2024semanticentropyprobesrobust} further proposed estimating “semantic entropy” from latent representations to enhance the robustness of uncertainty estimation.
	
	\begin{figure*}[h]
		\centering
		\includegraphics[width=0.95\linewidth]{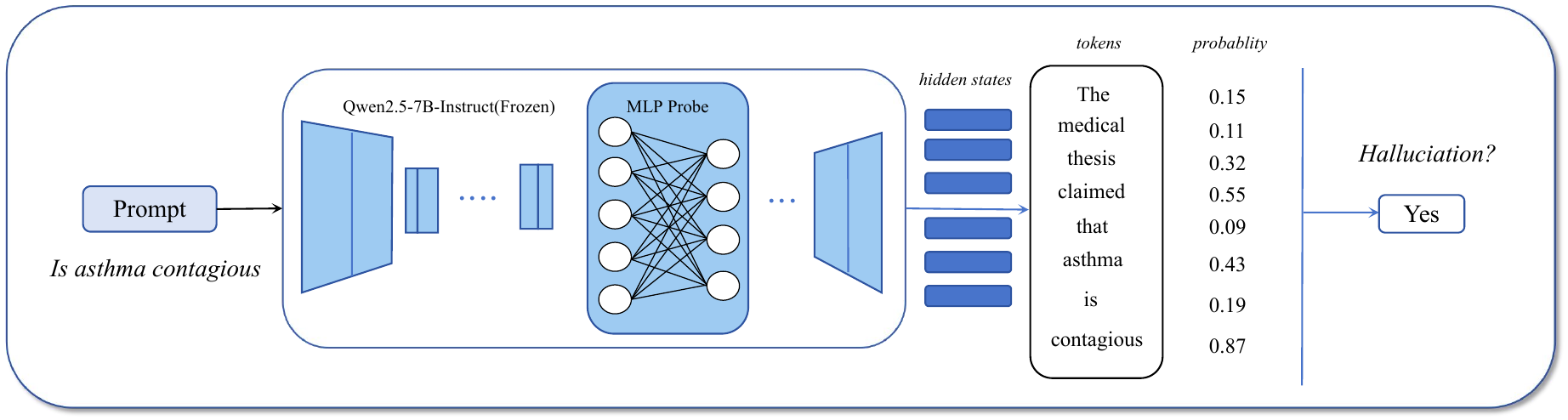}
		\caption{A General Framework for Language Model Hallucination Detection Based on Neural Network Probes. The model extracts latent states from frozen intermediate layers of Qwen2.5-7B-Instruct and computes hallucination probabilities for each token via MLP probes to enable real-time detection.}
		\label{fig:flow}
	\end{figure*}
	
	Knowledge-graph–based hallucination detection relies on external knowledge verification. Through retrieval-augmented generation or fact-checking, generated outputs are matched against external knowledge bases to assess factuality. ~\cite{lewis2021retrievalaugmentedgenerationknowledgeintensivenlp} first combined dense retrieval with seq2seq to inject external documents before generation, improving factual accuracy. Subsequent work ~\cite{shi2023replugretrievalaugmentedblackboxlanguage,gao2023rarrresearchingrevisinglanguage} enhanced factual alignment via plug-and-play retrieval or post-generation revision. ~\cite{abdi-etal-2025-hallurag} directly applied retrieval to hallucination detection by measuring conflicts between retrieved evidence and generated content. While effective at ensuring factual consistency, these methods suffer from high latency, strong dependence on knowledge-base coverage, and limited cross-domain robustness.
	
	Probe-based methods operate internally by training lightweight classifiers on intermediate hidden states. Linear probes were first introduced by ~\cite{alain2018understandingintermediatelayersusing}, showing that hidden layers encode separable semantic hierarchies, leading to widespread use in LLM interpretability. ~\cite{hewitt-manning-2019-structural} applied linear probes to syntactic structure detection, and ~\cite{obeso2025realtimedetectionhallucinatedentities} extended them to token-level hallucination detection. However, linear probes capture only shallow semantic relationships and struggle to model complex nonlinear dependencies.
	
	\subsection{Layer Position Probe Performance Model}
	Current research often assumes either “linear feature separability across layers” or “monotonic semantic enhancement with depth” ~\cite{alain2018understandingintermediatelayersusing,hewitt-manning-2019-structural}, and manually selects probe layers. ~\cite{tenney2019learncontextprobingsentence} systematically analyzed semantic encoding across layers via edge-probing tasks, but relying on accuracy or empirical selection may obscure biases. ~\cite{voita-titov-2020-information} introduced an information-theoretic measure to quantify inter-layer information. Building on this, we propose a distribution-based layer separability metric and a Bayesian optimization–based search to automatically locate optimal probe layers. We theoretically prove its optimality, and experiments show the method finds optimal positions efficiently with fewer iterations.

\section{Methodology}
	This section first proposes a probe training framework for token-level hallucination detection, encompassing problem formalization, MLP probe architecture, and multi-objective loss function design. Building on this basis, we present a mathematical model of layer-position probe performance and illustrate how Bayesian optimization can automatically choose the best probe layer.
	
\subsection{Probe Training}

    This subsection outlines the complete training process for probes. We first formalize the problem definition, then introduce the MLP probe architecture, and finally present a method for designing multi-objective loss functions.
        
	\begin{figure*}[h]
		\centering
		\includegraphics[width=0.95\linewidth]{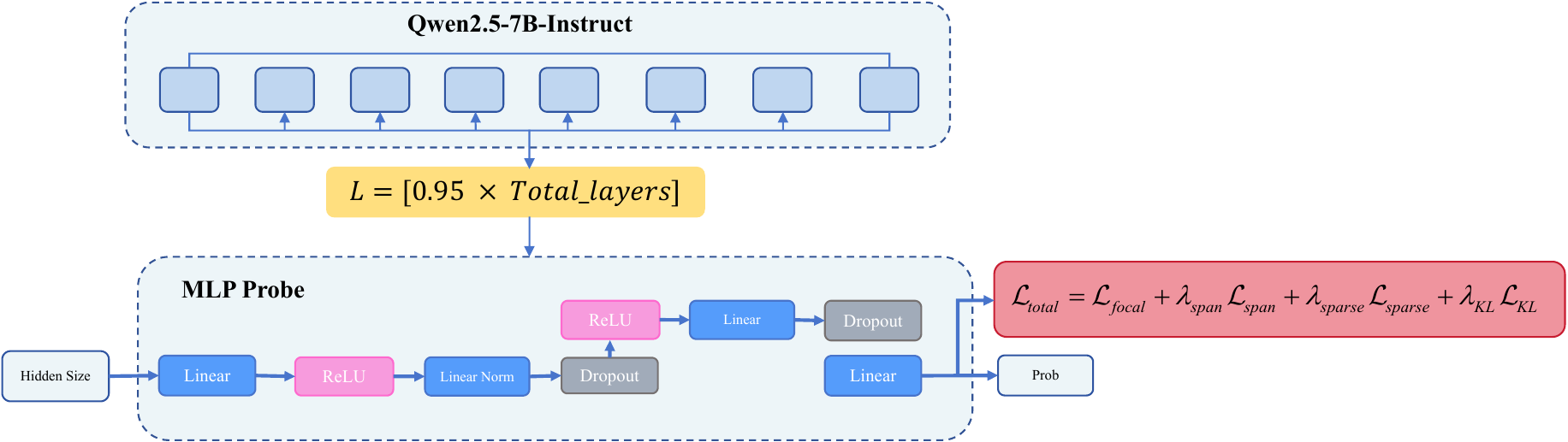}
		\caption{Multi-Layer Perceptron Probe Structure}
		\label{fig:netual}
	\end{figure*}
	
	\subsubsection{Problem Definition}
	
	Given a query $\bm{q}$ and a chat model $\bm{M}$, the model generates a sequence of tokens $t = (t_1, \ldots, t_n) \sim M(t|q)$. In the \emph{LongFact-annotations} dataset, each sample contains entity span annotations denoted as $s = [s^{start}, s^{end}]$, along with a binary label $y_s \in \{0, 1\}$. Here, $y_s = 1$ indicates that the entity span is hallucinated content. The hallucination detector aims to assign a probability to each annotated token $t_i$ in the generated sequence, representing the likelihood that this token belongs to hallucinated content. 
	Since hallucinated entities constitute an extremely small proportion of long texts and are interwoven with genuine content, given that hallucinations often stem from subtle shifts in the latent space, linear probes struggle to capture such nuanced variations. To address this, we propose a lightweight nonlinear MLP probe combined with a multi-objective loss function. This approach balances token-level accuracy, span-level consistency, sparsity, and model stability, thereby enhancing the reliability of hallucination detection.
	
	\subsubsection{MLP Probe}
	
	We propose a multilayer neural network probe, denoted as $\bm{M_{probe}}$, based on our original linear probe. This probe is attached to the base language model $\bm{M}$ and extracts hidden states from its $\ell$th intermediate layer as input. Compared to linear probes that contain only linear transformation numerical heads, the MLP probe employs a lightweight multi-layer perceptron to obtain token-level hallucination probabilities through multi-layer nonlinear mapping.
	
	\begin{equation}
		p_{i} = \mathrm{MLP}(h_i^{(\ell)})
	\end{equation}
	
	Here, $h_i^{(\ell)}$ denotes the hidden state of token $t_i$ in layer $\ell$. The MLP gradually compresses the feature dimensions using a number of layers of linear transformations, ReLU/GeLU activations, and normalization layers before producing a scalar probability value that indicates whether the token is part of a hallucination. Only the probe parameters are changed during training, and the base model $\bm{M}$ stays unchanged. Previous studies show that rich semantic structures and entity consistency properties are captured by probes at medium-to-high layers. Therefore, in order to take advantage of the high-level latent space with strong semantic abstraction capabilities, this work defaults to putting probes at $\ell = \lfloor 0.95 \times \text{num\_layers} \rfloor$.
	
	\subsubsection{Target Function}
	
		To achieve precise and stable probe training, we design the following multi-objective loss. By jointly optimizing local token discrimination and global span consistency, this loss formulation attains higher recall under low false positive rates and substantially improves AUC and convergence stability.
	
	For brevity, we outline only the core design; details are deferred to the appendix.
	
	\begin{equation}
		\mathcal{L}_{total} = 
		\mathcal{L}_{focal}
		+ \lambda_{span} \, \mathcal{L}_{span}
		+ \lambda_{sparse} \, \mathcal{L}_{sparse}
		+ \lambda_{KL} \, \mathcal{L}_{KL}
	\end{equation}%
	
	Specifically, $\mathcal{L}_{focal}$ dynamically modulates the cross-entropy between hallucinated and non-hallucinated tokens to amplify their dominant effects, thereby highlighting challenging samples and enhancing the model's sensitivity to hallucinated tokens. 
	In the meantime, we employ a Softmax-weighted span aggregation mechanism. For each sample $i$, we aggregate the positive and negative span sets $\mathcal{S}_{i}^{+}$ alongside $\mathcal{S}_{i}^{-}$ to compute the segment-level score $\hat{z_{i}}^{(s,e)}$ and average the span-level Focal loss, enhancing cross-token structural consistency and segment-level discriminative capability.
	 
	The sparse regularization term $\mathcal{L}_{sparse}$ limits probes to output high values only at a few important points, enhancing the accuracy of hallucination localization by preventing probe outputs from becoming too active across all tokens and losing interpretability. To further minimize noise interference, we use a high-loss masking technique based on the negative log-likelihood of the language model. A segment is designated as a high-loss area, and its labels are masked to improve training stability if its maximum negative log-likelihood satisfies $max(\mathcal{L}_{span}^{(i)}) > \tau$. Finally, to ensure the probe model with LoRA adapters maintains consistency with the original language model's distribution, we introduce the $\mathcal{L}_{KL}$ constraint for probe training. This enables non-intrusive training while preserving the original language modeling capabilities.

	\subsection{Layer-Probe Performance Modeling}
	In multi-layer Transformers, shallow layers lack semantic abstraction, while deep layers are shaped by language modeling and become task-agnostic. Thus, we propose a layer–probe performance model and use Bayesian optimization to adaptively locate the optimal probe layer. Specifically, we introduce the layer representation separability hypothesis and define a performance function to describe the interaction between feature separability, LoRA rank, and probe weights. We then analyze the coupling between probe optimization and language model loss to construct a joint utility function, theoretically explaining performance variations across probe layers. Finally, we present a Bayesian optimization-based method for automatically selecting optimal probe layers, enabling adaptive search for the best probe insertion positions.
	
	\subsubsection{Layered Separability Assumption}
	
	Assuming that the feature separability $S_{l}$ within each latent space layer can be characterized by the distributional difference between real and hallucinated samples, it is defined as:
	\begin{equation}
		S_l = D_{KL}\!\big(p(h_l|\text{real}) \,\|\, p(h_l|\text{hallucination})\big)
	\end{equation}
	Here, $D_{KL}(\cdot\|\cdot)$ denotes the KL divergence, which measures the representational difference between the two classes of samples at that layer.
	
	\subsubsection{Probe Performance Function}
	
	Probe performance $A(l,r,\alpha)$ and feature separability exhibit a positive correlation, but are constrained by LoRA capacity and loss weighting.
	Therefore, they can be approximated using a Sigmoid function as follows:
	\begin{equation}
		A(l,r,\alpha) = \sigma\!\big(\beta_0 + \beta_1 S_l - \gamma \frac{l}{Lr} - \eta \alpha \big)
	\end{equation}
	
	where $\sigma(x)$ denotes the sigmoid function.
	In the equation, $\beta_1 S_l$ represents the positive impact of separability enhancement on performance;
	$\frac{l}{Lr}$ signifies the conflict effect between high-level positions and low-rank capacity;
	$\eta \alpha$ indicates the disturbance to the main task distribution caused by excessively high probe weights.
	
	\subsubsection{Probe–Language Modeling Loss Coupling}
	
	Inserting probes disturbs the original distribution of the language model, so the language modeling loss can be expressed as:
	\begin{equation}
		\mathcal{L}_{lm}(l) = \mathcal{L}_{lm}^{0} + \lambda \alpha S_l
	\end{equation}
	Here, $\mathcal{L}_{lm}^{0}$ is the baseline language modeling loss of the frozen model, and $\lambda > 0$ controls how much probe optimization perturbs the main model.
	
	\subsubsection{Joint Optimization Objective}
	
	To simultaneously balance detection performance and language modeling stability, a normalized metric is defined:
	\begin{equation}
		\tilde{A}(l) = 
		\frac{A(l) - \min(A)}{\max(A) - \min(A)}
	\end{equation}
	\begin{equation}
		\tilde{\mathcal{L}}_{lm}(l) = 
		\frac{\mathcal{L}_{lm}(l) - \min(\mathcal{L}_{lm})}
		{\max(\mathcal{L}_{lm}) - \min(\mathcal{L}_{lm})}
	\end{equation}
	And construct a joint utility function:
	\begin{equation*}
		U(l) = w \cdot \tilde{A}(l) - (1-w) \cdot \tilde{\mathcal{L}}_{lm}(l),
	\end{equation*}
	Where $w \in [0,1]$ represents the performance-robustness trade-off coefficient, the optimal selection problem for probe layer positioning is modeled as:
	\begin{equation}
		l^* = \arg\max_{l \in [1, L]} U(l).
	\end{equation}
	
	\subsubsection{Bayesian Optimization Solution}
	
	\begin{algorithm}[h]
		\caption{Bayesian Optimization for Probe Layers}
		\label{alg:layer_probe}
		\begin{algorithmic}[1]
			\State \textbf{Input:} Transformer layers $L$, probe weight $\alpha$, LoRA rank $r$, robustness weight $w$, perturbation coefficient $\lambda$, maximum iteration count $T$
			\State \textbf{Output:} Optimal probe layer position $l^*$
			\Function{bayes}{$L$, $\alpha$, $r$, $w$, $\lambda$, $T$}
			\State Initialize $\mathcal{GP}$ surrogate model
			\State Construct the candidate layer set $CL$
			\State Randomly sample several initial layers for performance evaluation
			
			\For{Each initial layer $l$}
			\State $A[l] \leftarrow A_{\text{empirical}}(l)$
			\State $L_{lm}[l] \leftarrow L_{lm}^0 + \lambda \alpha S_l$
			\State $U[l] \leftarrow w \cdot \tilde{A}[l] - (1 - w)\cdot \tilde{L}_{lm}[l]$
			\State $\mathcal{GP} \leftarrow (l, U[l])$
			\EndFor
			
			\For{$t = 1$ to $T$}
			\State Select next layer via EI
			\State $l_{\text{next}} \leftarrow \arg\max_l \mathrm{EI}(l \mid \mathcal{GP})$
			
			\State Train probe at $l_{\text{next}}$ and evaluate $A[l_{\text{next}}]$
			\State $L_{lm}[l_{\text{next}}] \leftarrow L_{lm}^0 + \lambda \alpha S_{l_{\text{next}}}$
			\State $U[l_{\text{next}}] \leftarrow w \cdot \tilde{A}[l_{\text{next}}] - (1 - w) \cdot \tilde{L}_{lm}[l_{\text{next}}]$
			
			\State $\mathcal{GP} \leftarrow l_{\text{next}}, U[l_{\text{next}}$
			\EndFor
			
			\State $l^* \leftarrow \arg\max_{l \in CL} U[l]$
			
			\State \Return $l^*$
			\EndFunction
		\end{algorithmic}
	\end{algorithm}
	Since $U(l)$ lacks an analytical gradient and each evaluation requires a full train–validate cycle, we adopt Bayesian optimization with a Gaussian Process to search for the optimal layer $l^*$. For each candidate layer $l$, the main model is frozen while a probe is independently trained and evaluated using Recall@0.1 FPR on the validation set. The Bayesian optimizer treats this measured performance as a black-box response and builds a GP surrogate model:
	\begin{align}
		& \hat{U}(l) \sim \mathcal{GP}\big(m(l), k(l,l')\big) \\
		& l_{t+1} = \arg\max_{l} \mathrm{EI}(l \mid \hat{U})
	\end{align}
	Here, $\mathrm{EI}(\cdot)$ denotes the Expected Improvement criterion.
	The optimizer samples several initial layers, iteratively updates the surrogate model, selects the next candidate layer, and finally obtains the optimal position:
	\begin{equation}
		l^* = \arg\max_{l \in [1,L]} \mathrm{EI}(l).
	\end{equation}
	
	To justify the joint utility optimization problem, we prove the following theorem.
	
	\begin{figure*}[h]
		\centering
		\includegraphics[width=0.95\linewidth]{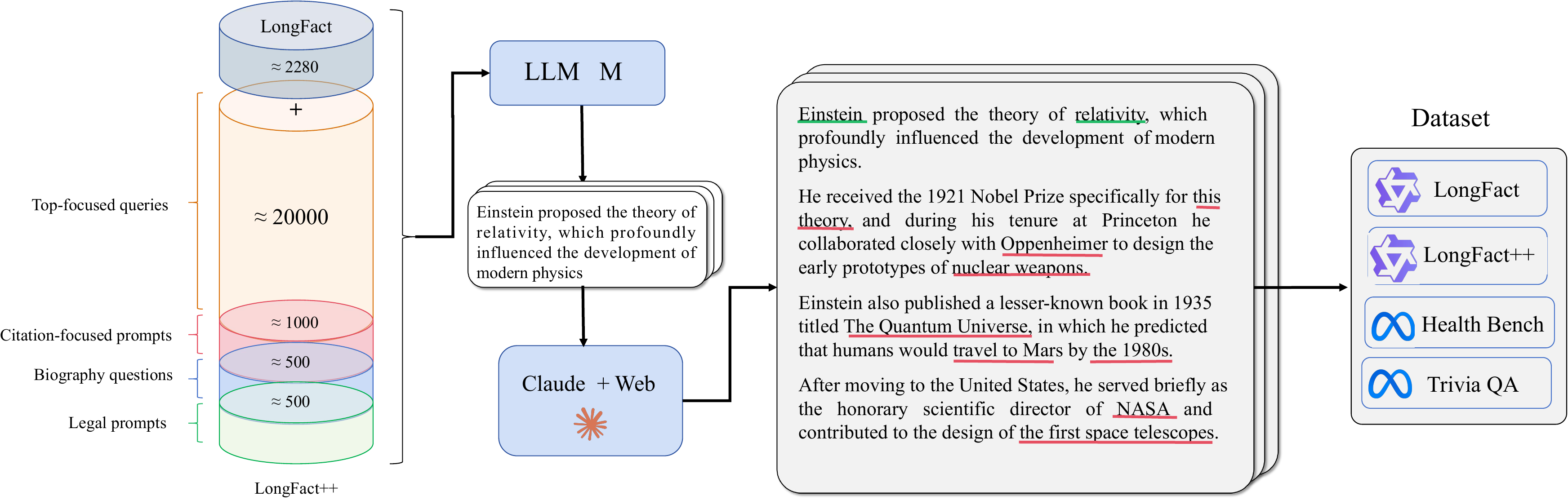}
		\label{fig:dataset}
		\caption{Entity-Level Token Annotation Process. Based on the \emph{LongFact++} prompt collection across several areas, the LLM produces long-form completions with both factual and imaginative content. After that, it uses an LLM with online search capabilities to find and verify entities in the created content, producing datasets from four different domains.}
	\end{figure*}
	
	\begin{table*}[h]
		\centering
		\small
		\begin{tabular}{llrrrrr}
			\toprule
			\textbf{Dataset} & \textbf{Method} &
			\textbf{AUC($\uparrow$)} & \textbf{R@0.1($\uparrow$)} &
			\textbf{Accuracy} & \textbf{Precision} & \textbf{Recall} \\
			\midrule
			
			\multirow{4}{*}{LongFact}
			& Perplexity & 0.4623 & 0.0879 & -- & -- & -- \\
			& Entropy    & 0.5124 & 0.1076 & -- & -- & -- \\
			& Linear Probe & 0.9022 & 0.6891 & 0.9022 & 0.8829 & 0.5280 \\
			& \textbf{MLP Probe} & \textbf{0.9528} & \textbf{0.7024} & \textbf{0.9528} & \textbf{0.8902} & \textbf{0.5715} \\
			\midrule
			
			\multirow{4}{*}{LongFact-Aug}
			& Perplexity & 0.4579 & 0.0873 & -- & -- & -- \\
			& Entropy    & 0.4963 & 0.1096 & -- & -- & -- \\
			& Linear Probe & 0.8894 & 0.6103 & 0.8894 & 0.3043 & 0.5799 \\
			& \textbf{MLP Probe} & \textbf{0.9404} & \textbf{0.6395} & \textbf{0.9404} & \textbf{0.4828} & \textbf{0.0784} \\
			\midrule
			
			\multirow{4}{*}{HealthBench}
			& Perplexity & 0.4605 & 0.0812 & -- & -- & -- \\
			& Entropy    & 0.4715 & 0.0948 & -- & -- & -- \\
			& Linear Probe & 0.9648 & 0.7350 & \textbf{0.9648} & 0.3962 & 0.3798 \\
			& \textbf{MLP Probe} & \textbf{0.9549} & \textbf{0.7695} & 0.9549 & \textbf{0.3231} & \textbf{0.4904} \\
			\midrule
			
			\multirow{4}{*}{TriviaQA}
			& Perplexity & 0.2996 & 0.0313 & -- & -- & -- \\
			& Entropy    & 0.3087 & 0.0301 & -- & -- & -- \\
			& Linear Probe & 0.8336 & 0.1730 & 0.8336 & 0.1178 & 0.3255 \\
			& \textbf{MLP Probe} & \textbf{0.9223} & \textbf{0.6891} & \textbf{0.9223} & \textbf{0.3886} & \textbf{0.4477} \\
			\bottomrule
		\end{tabular}
		\caption{
			Performance Comparison of Probes and Baseline Methods: Linear vs. MLP Probes
			Generalization Capability Across Different Datasets
			AUC and R@0.1 demonstrate the significant improvement of probes over baseline methods,
			while Accuracy/Precision/Recall illustrate the performance differences between MLP and linear probes.
		}
		\label{table:anlaysis}
	\end{table*}
	
	\begin{theorem}[Asymptotic Optimality of Bayesian Optimization]
		\label{thm:bo-optimality}
		Let the continuous bounded function $U(l)$ belong to an RKHS defined by the kernel $k(\cdot,\cdot)$, and assume that observations at layer $l$ are corrupted by sub-Gaussian noise with variance $\sigma^2$, i.e., $\tilde U(l) = U(l) + \varepsilon$. Using the GP-UCB strategy
		\begin{equation}
			l_t = \arg\max_{l \in \mathcal{L}} (\mu_{t-1}(l) + \sqrt{\beta_t} \sigma_{t-1}(l))
		\end{equation}
		where $\mu_{t-1}(l)$ and $\sigma_{t-1}(l)$ are the GP posterior mean and standard deviation, and $\beta_t$ is the confidence parameter, the cumulative regret after $T$ iterations satisfies
		\begin{equation}
			R_T = \sum_{t=1}^{T} \big(U(l^\ast) - U(l_t)\big) \le C_1 \sqrt{T \beta_T \gamma_T} + C_2
		\end{equation}
		with $\gamma_T$ denoting the information gain, and $C_1, C_2$ constants dependent on the noise variance and the RKHS norm bound. Consequently, the simple regret
		\begin{equation}
			r_T = \min_{t \le T} \big(U(l^\ast) - U(l_t)\big) \to 0, \quad T \to \infty
		\end{equation}
		indicating that the GP-UCB Bayesian optimization strategy asymptotically converges to the optimal layer $l^\ast$.
	\end{theorem}
	
	For the interest of space, we put the proof of this theorem to the appendix. 
	
	\section{Experiments}
	
	\subsection{Experiment Setup}
	
	\subsubsection{Baselines}
	To evaluate probe performance in contextual settings, we employ uncertainty-based metrics for comparison, specifically:
	
	\textbullet\ Token-level entropy: uncertainty of the next-token distribution.
	
	\textbullet\ Token-level perplexity: The uncertainty of the next token distribution.Higher values indicate the model considers many plausible continuations.
	
	\subsubsection{Model Training}
	To validate the effectiveness of our method, we train two types of probes—linear probes and MLP probes—on the Qwen2.5-7B-Instruct model. All probes are attached to the frozen base model without updating the original language model parameters.
	
	\subsubsection{Dataset}
	To train token-level hallucination detection probes, we use the \emph{LongFact-annotations} dataset, a long-text corpus constructed from prompts in \emph{LongFact} and \emph{LongFact++}. Unlike the \emph{SAFE framework}, which decomposes text into atomic claims and disrupts token alignment, this dataset preserves token-level sequences and focuses on entity categories such as persons, organizations, locations, dates, and citations. These factual entities can be verified against external sources while maintaining precise token boundaries. The prompts are divided into four main types: topic-focused queries, celebrity biography queries, citation-generation prompts, and milestone court case–related legal prompts.
	
	During annotation, the Claude 4 Sonnet model with web retrieval capabilities performs automatic fact verification. The system identifies entity spans, searches for supporting evidence, and labels each entity as “supported,” “unsupported,” or “insufficient information.” This process results in a high-quality, multi-domain dataset with precise token boundaries, providing a reliable foundation for training probe models.
	
	Experiments are conducted on four representative datasets to evaluate model robustness across different knowledge types and hallucination patterns.
	\begin{enumerate}
		\item \emph{LongFact} dataset generated by Qwen2.5-7B-Instruct;
		\item \emph{LongFact-Augmented} dataset generated by Qwen2.5-7B-Instruct;
		\item \emph{HealthBench} medical knowledge question-answering dataset generated by Meta-Llama-3.1-8B-Instruct;
		\item \emph{TriviaQA} Common-Sense Question-Answering Dataset generated by Meta-Llama-3.1-8B-Instruct.
	\end{enumerate}
	
	\subsubsection{Evaluation Metrics}
	The experiment employs metrics such as Accuracy, Precision, Recall, and R@0.1 at the token level to quantitatively evaluate the model's performance in the probe detection task.
	
	\subsection{Experimental Results}
	
	Table~\ref{table:anlaysis} compares the performance across different datasets between two types of uncertainty baselines—model perplexity and semantic entropy—and the linear probe and MLP probe in terms of AUC and R@0.1. It also compares the generalization ability between the linear probe and the MLP probe. Overall, the performance of the perplexity and semantic-entropy-based methods is relatively poor across all datasets, indicating that relying solely on the uncertainty generated by the language model itself is insufficient to reliably capture token-level hallucinations.
	
	\begin{figure*}[t]
		\centering
		\begin{subfigure}{0.34\textwidth}
			\centering
			\includegraphics[width=\linewidth]{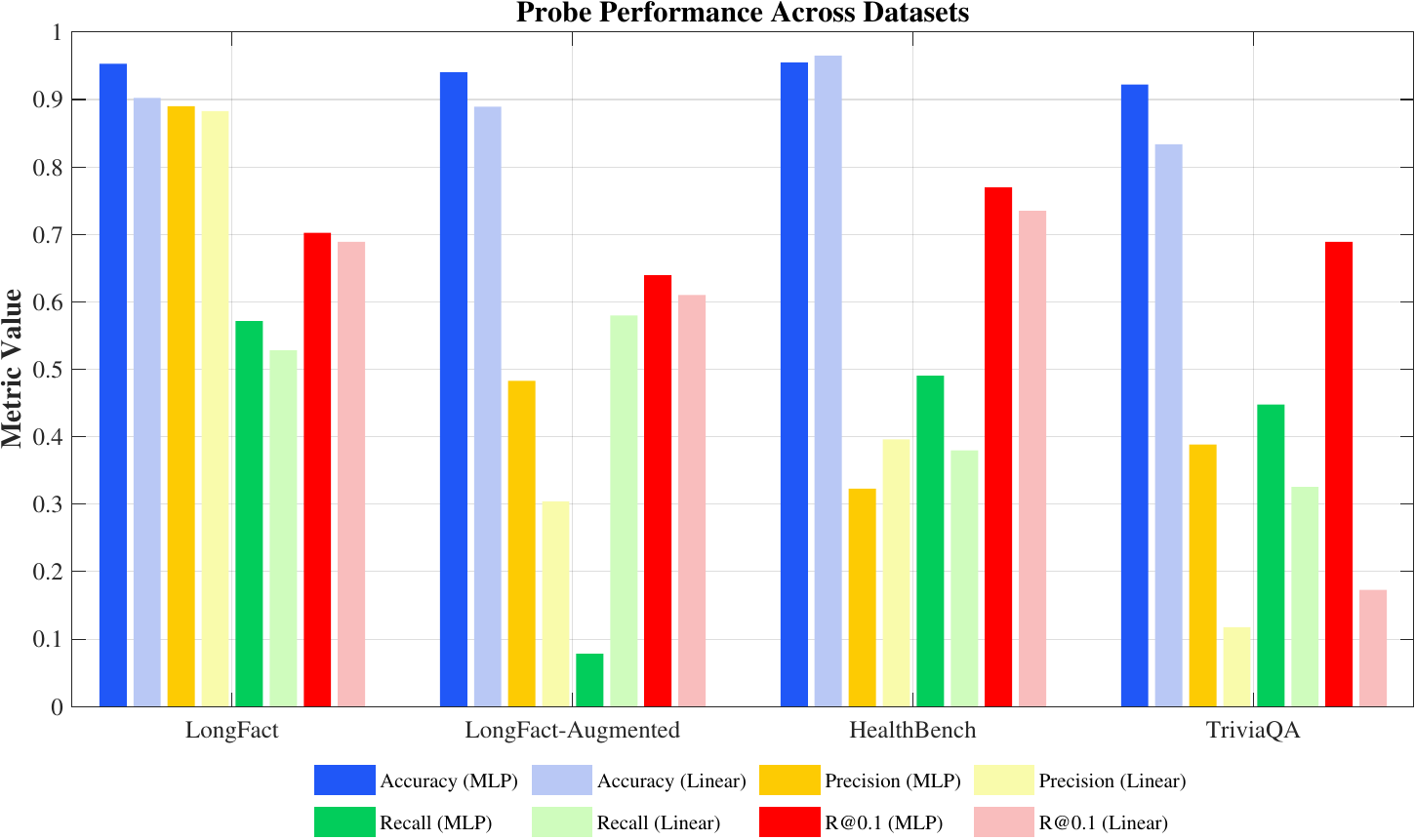}
			\caption{Performance comparison on the dataset}
			\label{fig:contrast}
		\end{subfigure}
		\hfill
		\begin{subfigure}{0.3\textwidth}
			\centering
			\includegraphics[width=0.97\linewidth]{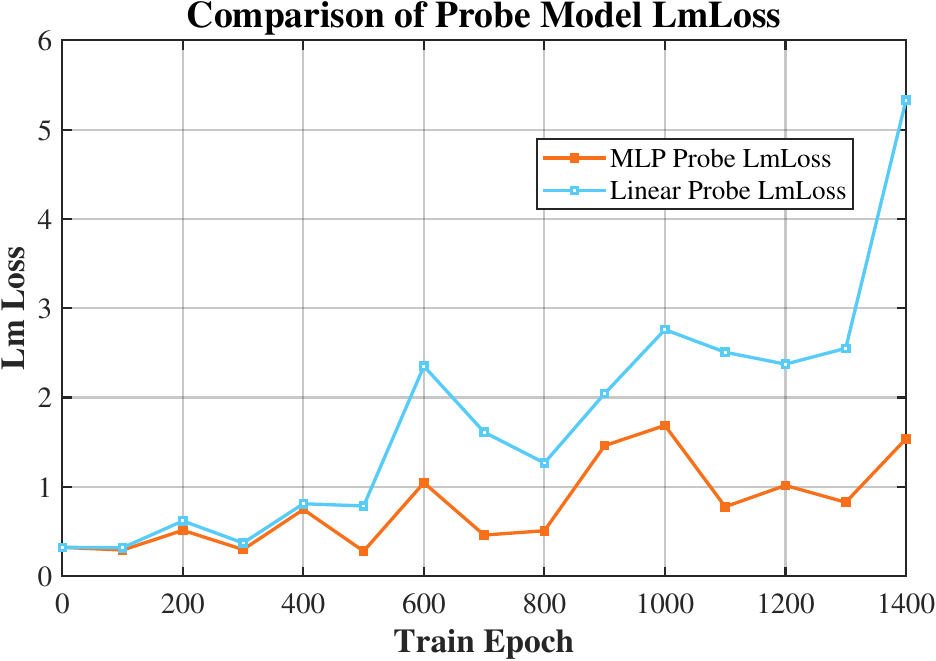}
			\caption{Language modeling loss}
			\label{fig:lm_loss}
		\end{subfigure}
		\hfill
		\begin{subfigure}{0.30\textwidth}
			\centering
			\includegraphics[width=0.98\linewidth]{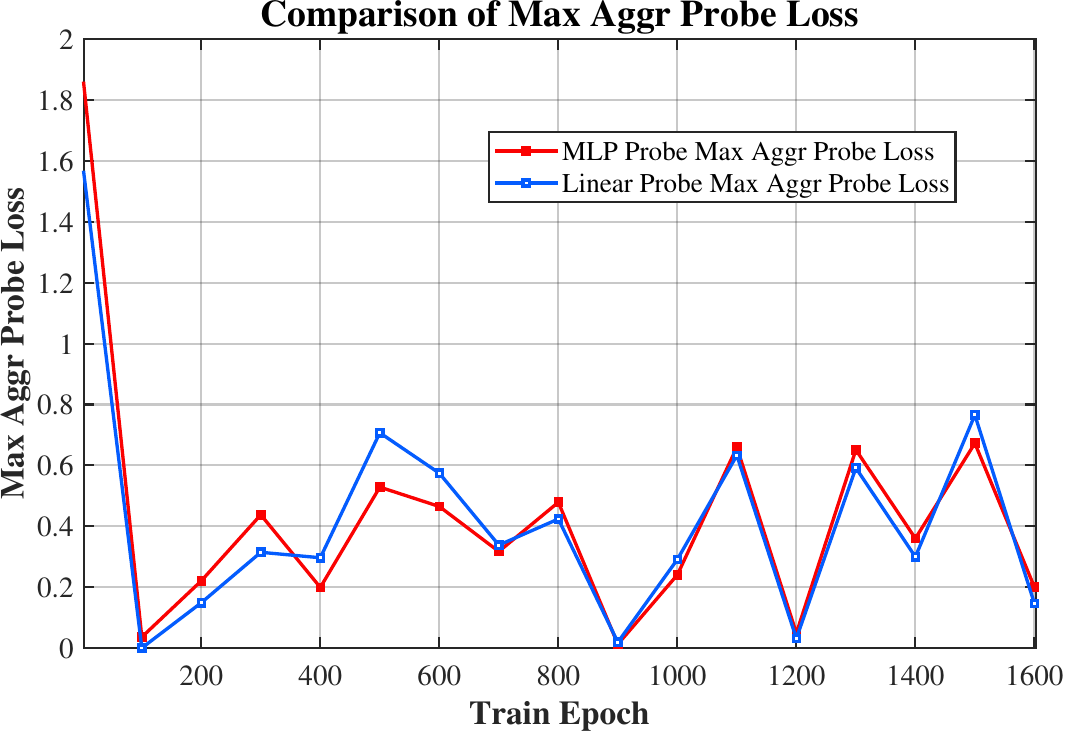}
			\caption{Aggregated label prediction loss}
			\label{fig:aggr_loss}
		\end{subfigure}
		\vspace{-0.5em}
		\caption{MLP probes versus linear probes across different tasks. Figures (a)–(c) correspond to the dataset performance, language modeling loss, and aggregated label prediction loss, respectively.}
		\label{fig:probe_all}
	\end{figure*}

	\figurename~\ref{fig:contrast} demonstrates that MLP probes consistently outperform linear probes on the factual datasets \emph{LongFact} and \emph{LongFact-Augmented}, achieving a 5.605\% accuracy improvement and 8.2\% recall on \emph{LongFact}. illustrating that the MLP probe captures nonlinear semantic features more effectively in intermediate-layer hidden states, thereby enhancing sensitivity to potential hallucination samples. Furthermore, on the \emph{HealthBench} and \emph{TriviaQA} datasets, the MLP probe demonstrated stronger generalization capabilities: achieving higher recall and R@0.1 than the linear probe on the medical question-answering dataset, while achieving over 270\% improvement in Precision on the common-sense question-answering dataset. while Recall increased by 37\%, indicating its enhanced adaptability to complex, cross-domain linguistic features.
	
	\figurename~\ref{fig:lm_loss} and \figurename~\ref{fig:aggr_loss} indicate that the improvement of MLP probes over linear probes in terms of Probe Loss is relatively limited, meaning both exhibit similar feature fitting capabilities at the probe layer. However, in terms of LM Loss, the MLP probe significantly outperforms the linear probe, achieving the same convergence effect in substantially less training time. This demonstrates its superior expressive capability in capturing underlying semantic information and reducing language generation errors.
	
	\begin{table}[h]
		\centering
		\begin{tabular}{lcc}
			\toprule
			\textbf{Component} & $\Delta$ \textbf{AUC}($\downarrow$) & \textbf{R@0.1}($\downarrow$)\\
			\midrule
			$\mathcal{L}_{focal}$ & -0.3720 & -0.4414 \\
			$\mathcal{L}_{sparse}$& -0.0567 & -0.1393 \\
			$\mathcal{L}_{KL}$    & -0.0234 & -0.0552 \\
			$\mathcal{L}_{span}$  & -0.0132 & -0.1706 \\
			\bottomrule
		\end{tabular}
		\caption{Analysis of the contribution of each module in the joint loss function to probe performance. $\Delta$ AUC represents the performance degradation after removing the module.}
		\label{tab:ablation2}
	\end{table}
	
	Table~\ref{tab:ablation2} shows the impact of different components in the joint loss function on probe performance. Removing $\mathcal{L}_{focal}$ leads to a significant drop in probe performance, highlighting its central role in mitigating class imbalance and enhancing discrimination of hallucinated tokens.And the removal of $\mathcal{L}_{sparse}$, $\mathcal{L}_{KL}$, or $\mathcal{L}_{span}$ results in mild performance degradation. Notably, $\mathcal{L}_{span}$ has a more pronounced effect on R@0.1, indicating that it improves early recall by enhancing sensitivity to consecutive hallucinated spans. In contrast, $\mathcal{L}_{sparse}$ primarily affects AUC, suggesting its role in suppressing over-activation and constraining redundant signals across tokens.
		
	Overall, the ablation study validates the rationale behind our loss design: $\mathcal{L}_{focal}$ provides the main discriminative gain, while $\mathcal{L}_{span}$ and $\mathcal{L}_{sparse}$ offer complementary benefits via local structure modeling and representation regularization, and $\mathcal{L}_{KL}$ yields additional improvements in edge cases. The results reflect the synergistic effect of the multi-objective loss in capturing token-level hallucination features.
	
	\begin{figure}[htbp]
		\centering
		\includegraphics[width=0.95\linewidth]{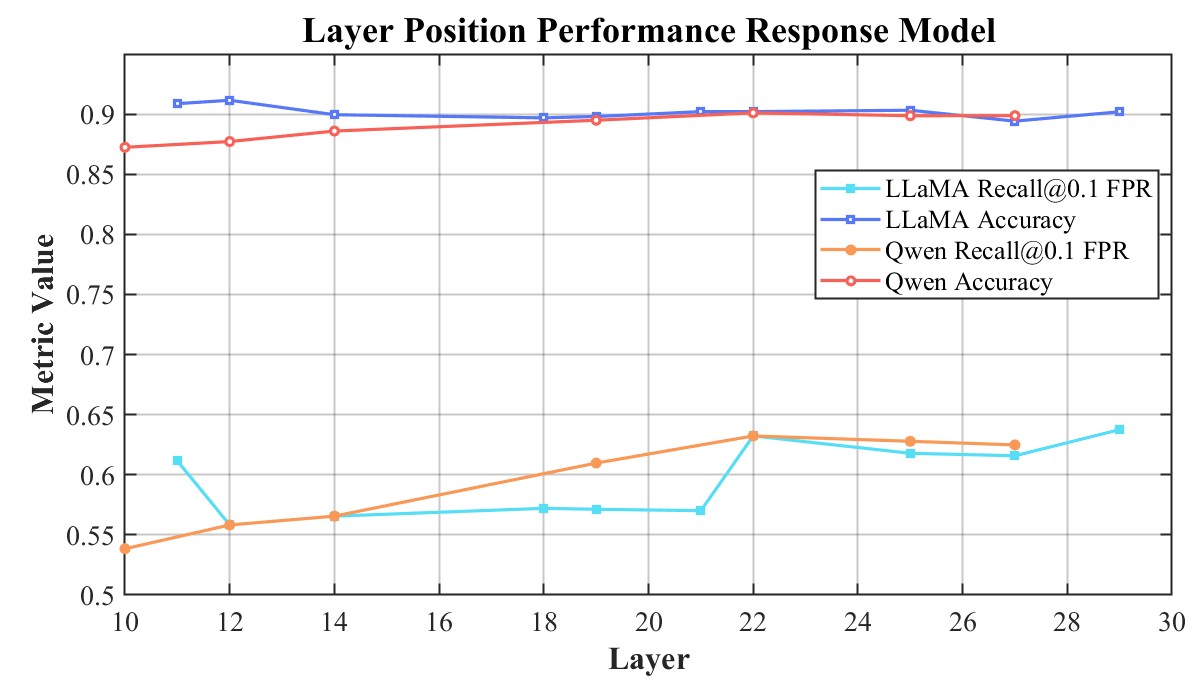}
		\caption{The established layer position probe performance model iteratively searches for optimal layer positions in Qwen2.5-7B-Instruct and Meta-Llama-3.1-8B-Instruct. It can be observed that the optimal layer position for the Qwen2.5-7B-Instruct model is at layer 29, while the optimal layer position for the Meta-Llama-3.1-8B-Instruct model is at layer 22.}
		\label{fig:model}
	\end{figure}
	
	\figurename ~\ref{fig:model} illustrates the correspondence between different layer positions $l$ and the metric R@0.1 in the Bayesian optimization search process, as established by the mathematical model of “layer position–probe performance” developed in this paper. The search space was defined as $l \in [1,L]$, with convergence achieved through 3 random initializations and 5 Bayesian iteration updates.
	
	Through joint optimization of the Gaussian process surrogate model and the expected improvement criterion, the optimal layer position was ultimately determined to be layer 22, where the probe performance achieved a global optimum. Results demonstrate that the model exhibits optimal feature separability and hallucination detection capabilities in the mid-to-high layer semantic representation region.
	
	\section{Conclusion}
	
	We propose a neural network probe–based token-level hallucination detection framework that uses lightweight supervised learning on LLM hidden states to identify hallucinated tokens in real time. By employing a multi-objective loss for MLP probes and modeling the relationship between probe placement and performance, our method offers an efficient, scalable, and interpretable approach for internal LLM analysis. While challenges remain, it provides a foundation for improving token-level output validity. Future work may extend detection beyond entities, explore cross-lingual and multimodal applications, and combine with retrieval or external knowledge methods to enhance robustness.
	
	\bibliographystyle{named}
	\bibliography{ijcai26}
	
\end{document}